\documentclass[twocolumn,3p]{elsarticle}
\usepackage[T1]{fontenc}

\usepackage[utf8]{inputenc}
\usepackage{graphicx}
\usepackage{epstopdf}
\usepackage{caption}
\usepackage{subcaption}
\usepackage{amsmath}
\usepackage{amsfonts}
\usepackage{amssymb}
\usepackage{amsthm}
\usepackage{float}
\usepackage{hyperref}
\usepackage{enumitem}

\usepackage{pgfplots}

\usepackage{booktabs}

\newdefinition{rmk}{Remark}
\newproof{pf}{Proof}

\journal{peer review}

\begin{document}

\begin{frontmatter}

\title{A Surveillance Game between a Differential Drive Robot and an Omnidirectional Agent: The Case of a Faster Evader}

\author[1]{Rodrigo Saavedra}
\ead{rsaavedra@cicese.mx}
\author[1]{Ubaldo Ruiz\corref{cor1}}
\ead{uruiz@cicese.mx}

\affiliation[1]{organization={Centro de Investigaci\'on Cient\'ifica y de Educaci\'on Superior de Ensenada},
addressline={Carretera Ensenada - Tijuana No. 3918, Zona Playitas},
postcode={22860},
city={Ensenada},
country={M\'exico}}

\cortext[cor1]{Corresponding author}


\begin{abstract}
A fundamental task in mobile robotics is to keep an agent under surveillance using an autonomous robotic platform equipped with a sensing device. Using differential game theory, we study a particular setup of the previous problem. A Differential Drive Robot (DDR) equipped with a bounded range sensor wants to keep surveillance of an Omnidirectional Agent (OA). The goal of the DDR is to maintain the OA inside its detection region for as much time as possible, while the OA, having the opposite goal, wants to leave the regions as soon as possible. We formulate the problem as a zero-sum differential game, and we compute the time-optimal motion strategies of the players to achieve their goals. We focus on the case where the OA is faster than the DDR.  Given the OA's speed advantage, a winning strategy for the OA is always moving radially outwards to the DDR's position. However, this work shows that even though the previous strategy could be optimal in some cases, more complex motion strategies emerge based on the players' speed ratio. In particular, we exhibit that four classes of singular surfaces may appear in this game: Dispersal, Transition, Universal, and Focal surfaces. Each one of those surfaces implies a particular motion strategy for the players.
\end{abstract}

\end{frontmatter}

\section{Introduction}
\label{sec:introduction}

Many tasks in mobile robotics can be viewed as pursuit-evasion games. Some examples include a convoy of autonomous vehicles following a leader, a robot guard trying to capture a suspicious agent, or a group of robotic routers navigating an environment to establish connectivity with an agent. A fundamental task in mobile robotics is to keep an agent under surveillance using an autonomous robotic platform equipped with a sensing device. Using differential game theory, we study a particular setup of the previous problem. A Differential Drive Robot (DDR) equipped with a bounded range sensor wants to keep surveillance of an Omnidirectional Agent (OA) as it moves in an environment without obstacles. The goal of the DDR is to maintain the OA inside its detection region for as much time as possible, while the OA, having the opposite goal, wants to leave the regions as soon as possible. The DDR is subject to non-holonomic constraints, and it can only change its motion direction at a bounded rate that is inversely proportional to its translational speed, which is also bounded. On the other hand, the OA can instantaneously change its motion direction and only has a bound in its maximum speed. We formulate the problem as a zero-sum differential game, and we compute the time-optimal motion strategies of the players to achieve their goals. A version of this problem \cite{RUIZ-22}, in which the OA is slower than the DDR, has been addressed in the past; however, this current work focuses on the case where the OA is faster than the DDR. Given the OA's speed advantage, one may think the OA's strategy to escape is always moving radially outwards to the DDR's position. However, we show that a more complex set of motion strategies emerges based on the players' speed ratio. In particular, we exhibit that four classes of singular surfaces may appear in this version of the game: Dispersal, Transition, Universal, and Focal surfaces. Two of them, Universal and Focal surfaces, are absent in the case of a slower OA studied in \cite{RUIZ-22}.

This work employs R. Isaacs' methodology \citep{ISAACS-65, BASAR-98, LEWIN-12} for analyzing and solving differential games. The methodology focuses on solving boundary value problems through constrained optimization. Initially, we compute the game's final configurations, where the evader escapes despite the pursuer's efforts. With this information, we perform a backward integration of the motion equations, starting from these final configurations and considering the players' optimal controls. This process helps determine the trajectories that lead to the game's termination, specifically the evader's escape. The integration is conducted to ensure the computed trajectories minimize a cost function, which, in this case, is the time to escape.

The methodology involves dividing the playing space into regions where the value function is differentiable. Within each region, specific motion strategies for the players are computed. The most challenging aspect of this process typically involves identifying the boundaries of these regions, known as "singular surfaces." On a singular surface \citep{BASAR-98}, one of the following conditions holds: 1) the players' strategies are not uniquely defined, 2) the value function is not continuously differentiable, or 3) the value function is discontinuous. J. Lewin thoroughly examined singular surfaces and methods for their computation in \cite{LEWIN-12}.

When solving a differential game, it is generally unknown in advance whether the optimal trajectories will include singular portions. The existence of a singular surface and its associated tributary trajectories only becomes apparent if the regular backward construction of candidate trajectories fails to cover the playing space. This situation can be observed in our game after computing the primary solution \cite{LEWIN-12}.

In most cases, to realize a singular surface and its corresponding outcome, one player must base their choice of controls on prior knowledge of their opponent's control choice. A strategy computed using this information is considered a non-admissible strategy in differential game theory. In contrast, an admissible strategy relies solely on the knowledge of the system's state and does not require additional information about the players' controls. In this work, we characterize the singular surfaces that appear in the solution of our problem and the trajectories that fill the regions defined by them. The trajectories within a region correspond to admissible strategies for the players.

Similar to other previous works \cite{RUIZ-13, RUIZ-22}, we have successfully derived analytical expressions that describe the players' motion strategies in this work. 

\subsection{Previous Work}
\label{sec:prev}

The literature about pursuit-evasion games is vast; however, one can distinguish three main classes of them: search, capture, and tracking. Search problems aim to find one evader agent that moves in an environment using another pursuer agent. In capture games, the goal is to reach a certain distance to the evader, usually minimizing a given cost function, such as time. Finally, in tracking problems, the goal is to keep surveillance of an evader as it moves. The previous problems can be extended to consider several pursuers and evaders. For a more detailed taxonomy of pursuit-evasion problems, we suggest to the reader the following surveys \citep{CHUNG-11,ROBIN-16}. In the following paragraphs, we describe those works we consider to be the closest ones to our problem.

In \cite{RUIZ-13}, the problem of capturing an OA using a DDR in minimum time is addressed. Like our current work, the OA plays as an evader, and the DDR is a pursuer. The problem is framed as a zero-sum differential game, and its solution comprises the computation of Universal, Dispersal, and Transition Surfaces. Our work varies from \cite{RUIZ-13} in two crucial aspects. First, the players' goals differ; in our case, the pursuer wants to keep the OA inside its detection region for as long as possible, while in \cite{RUIZ-13}, the DDR wants to capture the OA by reaching a given distance. Analogous, in our case, the OA wants to escape as soon as possible, while in \cite{RUIZ-13}, the OA wants to delay the capture. That implies that the players' motion strategies computed in \cite{RUIZ-13} cannot be used in our case. Second, a Focal Surface appears in our current work, which is not present in \cite{RUIZ-13}.

Another work addressing a pursuit-evasion game between an OA and a DDR was presented in \cite{RUIZ-23}. The problem of an OA evading surveillance in minimum time from a DDR equipped with a limited field-of-view sensor is studied. The detection region is modeled as a semi-infinite cone. In \cite{RUIZ-23}, the authors find the time-optimal motion strategies of the players to achieve their goals. In particular, they exhibited the existence of three classes of singular surfaces in the game's solution: Dispersal, State constraint, and Equivocal Surfaces. One main difference between \cite{RUIZ-23} and our current work is that in our game, the evader escapes by increasing the distance to the pursuer, while in \cite{RUIZ-23}, it does by leaving the sides of the cone. This change in the game's terminal condition implies that our game cannot be solved by employing the players' motion strategies found in \cite{RUIZ-23}.

The most closely related work to our current problem was presented is \cite{RUIZ-22}. In that work, the problem of keeping surveillance of an OA with a DDR equipped with a range sensor is studied for the first time. In particular, a version in which the OA is slower than the DDR was addressed. Our current work focuses on the case where the OA is faster than the DDR. Given the OA's speed advantage, one may think the OA's strategy to escape is always moving radially outwards to the DDR's position, and the solution in the entire playing space should be simpler than the one found in \cite{RUIZ-22}. However, we show that a more complex set of motion strategies than the ones in \cite{RUIZ-22} emerges based on the players' speed ratio. In particular, we exhibit that four classes of singular surfaces may appear in this version of the game: Dispersal, Transition, Universal, and Focal surfaces. Two of them, Universal and Focal surfaces, are absent in the case of a slower OA studied in \cite{RUIZ-22}.

\subsection{Main contributions}

The main contributions of this work are the following : 
\begin{enumerate}
    \item We compute closed-form expressions for the time-optimal motion strategies of the players.
    \item We exhibit the existence of four classes of singular surfaces that may appear in this game: Dispersal, Transition, Universal, and Focal surfaces.
    \item We characterize the game's solution based on the players' speed.
    \item We present numerical simulations to illustrate the players' time-optimal motion strategies.
\end{enumerate}

\section{Problem definition}
\label{sec:problem_formulation}

\begin{figure*}[t!]
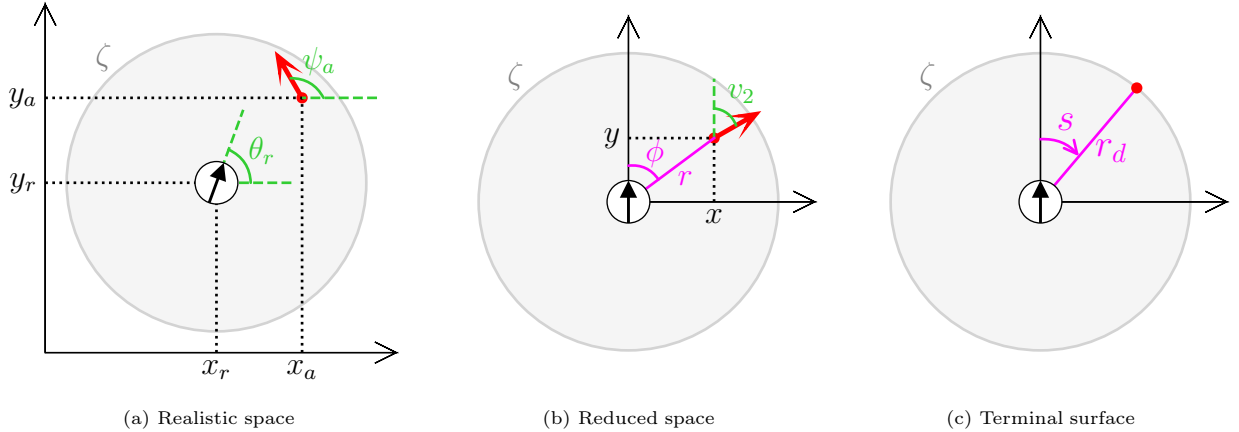

    \centering
    \subfloat[Realistic space]{
        \input{fig1_Sketch_a.pgf}
        \label{fig:realistic}
    }%
    \subfloat[Reduced space]{
        \input{fig1_Sketch_b.pgf}
        \label{fig:reduced}
    }%
    \subfloat[Terminal surface]{
        \input{fig1_Sketch_c.pgf}
        \label{fig:terminal}
    }%
    \caption{%
        The pursuer DDR is represented by the white disc of radius~$b$. %
        The evader OA is represented by the red dot. %
        The larger gray circle in the background represents the detection region of radius~$r_d$. %
    }
\end{figure*}

An Omnidirectional Agent (OA) and a Differential Drive Robot (DDR) move in the Euclidean plane. The DDR is equipped with a range sensor modeled as a circle of radius $r_d$. We study a pursuit-evasion problem in which the DDR (pursuer) wants to maintain the OA (evader) inside its detection region for as long as possible. On the contrary, the OA seeks to escape it as soon as possible. We formulate the problem as a zero-sum differential game that ends when the OA reaches a distance $r_d$ from the DDR despite any resistance of this player. The players have bounded maximum speeds $V_r^{\max}$ for the DDR and $V_a^{\max}$ for the OA. Different from \cite{RUIZ-22}, we assume that the OA is faster than the DDR, i.e., $V_a^{\max}>V_r^{\max}$. We focus on a purely kinematic problem and do not consider any effects due to dynamic constraints. We represent the pose of the DDR as $(x_r,y_r,\theta_r)$, and its motion is described by the following equations \cite{BALKCOM-02}
\begin{equation}
\label{eq:DDRmotion}
\begin{split}
\dot x_r &= \left(\frac{u_1 + u_2 }{2}\right) \cos \theta_r, \:\:
\dot y_r = \left(\frac{u_1 + u_2 }{2}\right) \sin \theta_r, \\
\dot \theta_r &= \left(\frac{u_2  - u_1 }{2b}\right), \\
\end{split}
\end{equation}
where $u_1,u_2 \in [-V_r^{\max},V_r^{\max}]$ are the controls (velocities) of the left and right wheels. $b$ is the distance between the DDR's center and the wheel's location. For a DDR, we have that there is an inverse relation between the translational and rotational velocities, which is described by
\begin{equation}
\label{eq:DDRrelation}
|\dot \theta_r| = \frac{1}{b}\left|V_r^{\max}-\left(\frac{u_1+u_2}{2}\right)\right|.
\end{equation}
The position of the OA is represented by $(x_a, y_a)$ and its motion is described by
\begin{equation}
\label{eq:OAmotion}
\dot x_a = v_1 \cos \psi_a, \:\: \dot y_a = v_1 \sin \psi_a,
\end{equation}
 where $v_1\in[0,V_a^{\max}]$ is the evader's speed and $\psi_a\in[0,2\pi)$ is its motion direction. The state of the system can be denoted as $\mathbf{x}=(x_r,y_r,\theta_r,x_a,y_a)\in \mathbb{R}^2\times S^1 \times \mathbb{R}^2$. In this work, the previous representation is called the {\em realistic space}, and all angles are measured in a counter-clockwise direction (see Fig. \ref{fig:realistic}). 
For solving pursuit-evasion games, one usually uses a reference frame fixed to one of the players to simplify the computations.

In this work, we employ a reference frame fixed to the DDR's body, where the $y$-axis is aligned to the DDR's heading. We call this representation the {\em reduced space}. The system's state (evader's relative position) can be expressed as $\mathbf{x}_R = (x, y) \in \mathbb{R}^2$. In the reduced space, all the orientations are measured with respect to the positive $y$-axis in a clockwise sense (see Fig. \ref{fig:reduced}). The following coordinate transformations relate the reduced and realistic spaces
\begin{equation}
\label{eq:transformation}
\begin{split}
x &= (x_a-x_r) \sin \theta_r - (y_a - y_r) \cos \theta_r, \\
y &= (x_a-x_r) \cos \theta_r + (y_a - y_r) \sin \theta_r, \\
v_2 &= \theta_r - \psi_a,
\end{split}
\end{equation}
where $v_2$ denotes the OA's motion direction in the new reference frame. Computing the time derivatives of Eq. (\ref{eq:transformation}), and substituting Eqs. (\ref{eq:DDRmotion}) and (\ref{eq:OAmotion}) into the resulting expresions, we obtain the motion equations in the reduced space
\begin{equation}
\label{eq:reducedsystem}
\begin{split}
&\dot x =\left(\frac{u_2  - u_1 }{2b}\right)y + v_1 \sin v_2, \\
&\dot y =-\left(\frac{u_2  - u_1 }{2b}\right)x - \left(\frac{u_1 + u_2 }{2}\right) + v_1 \cos v_2.\\
\end{split}
\end{equation}
We have that $\dot{\mathbf{x}}_R=f(\mathbf{x}_R,u,v)$ where $u=(u_1,u_2)\in[-V^{\max},V^{\max}]\times[-V^{\max},V^{\max}]$ and $v=(v_1,v_2)\in[0,V_a^{\max}]\times[0,2\pi)$.
In this game, having a polar representation of the reduced space is also convenient. In polar coordinates, the system's state is represented by $\mathbf{x}_P=(r,\phi)$, where $r$ denotes the length of the segment joining the frame's origin and the evader's location, and $\phi$ is its orientation. The motion equations in polar coordinates are given by
\begin{equation}
\label{eq:polarsystem}
\begin{split}
\dot r &= v_1 \cos(v_2 - \phi) - \left(\frac{u_1+u_2}{2}\right) \cos \phi, \\
\dot \phi &= \left( \frac{u_2-u_1}{2b} \right) + \frac{v_1\sin(v_2-\phi)}{r} + \left(\frac{u_1+u_2}{2}\right)\frac{\sin \phi}{r}.
\end{split}
\end{equation}
We switch between both representations since part of the analysis, and computations are easier to perform in one representation than in the other. For the remainder of the paper, we introduce two useful definitions. First, $\rho_{v}=V_a^{\max}/V_r^{\max}$ represents the ratio between the maximum translational speeds of the OA and the DDR. Second, $\rho_d=b/r_d$ represents the ratio between the distance from the robot's center to the wheel's location $b$ and the radius $r_d$ of the pursuer's detection region.

\section{Overview of the solution}
\label{sec:overview}

\begin{figure*}[t!]
    \centering
    \input{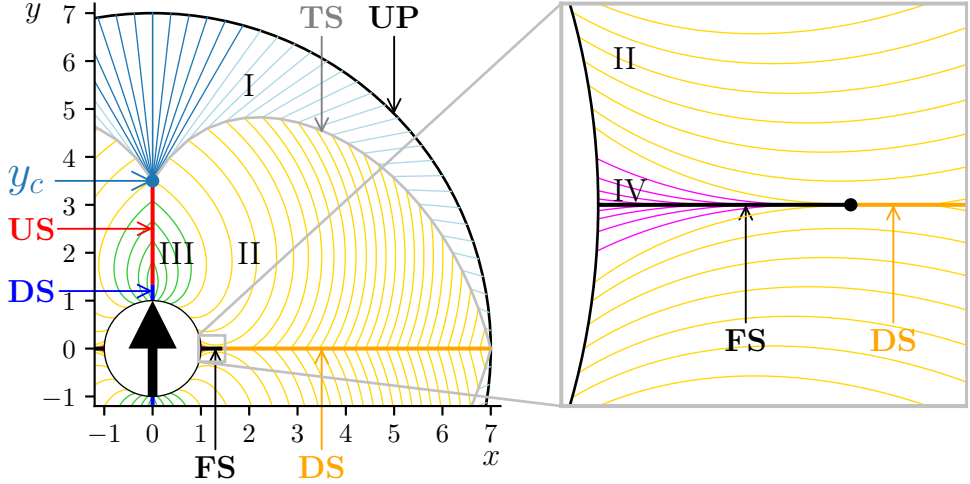}
    \caption{
        Partition of the playing space in reduced coordinates. %
        The white disk represents the DDR and the large arrow its direction of motion. %
        The outer circle corresponds to the Usable Part (UP), the silver line to the Transition Surface (TS), the vertical red line to the Universal Surface (US), blue vertical and orange horizontal lines are Dispersal Surfaces (DS). %
        The zoom-in shows tributaries that emanate from the Focal Surface (FS). %
    }
    \label{fig:partition}
\end{figure*}

In this section, we provide an outline of the problem's solution to guide the identification and computation of the players' motion strategies later in the paper. A partition of the playing space for $V_r^{\max}=1m/s$, $V_a^{\max}=2m/s$, $b=1m$ and $r_d=7m$ is shown in Fig. \ref{fig:partition}. That figure exhibits four classes of singular surfaces that may appear in this game: Dispersal (DS), Transition (TS), Universal (US), and Focal (FS). The partition of the playing space is based on the tributary trajectories reaching each class of singular surface. 

The set of trajectories in each region of the reduced space are as follows:
\begin{enumerate}[label=\Roman*.,itemsep=0pt]
    \item Tributaries of the UP (primary solution) described by a straight-line given by Eq. (\ref{eq:straightline}), they reach the TS when $s>\arctan(\rho_d\rho_v)$, otherwise they converge at the critical point $(0,y_c)=(0,r_d/\rho_v)$. 
    \item Tributaries of the TS given by Eq. (\ref{eq:rotation}), they reach either the horizontal DS at $y=0$ or the DDR's body at $r=b$. 
    \item Tributaries of the US given by Eq. (\ref{eq:rotationUS}), they reach either the vertical DS at $x=0$ or the DDR's body at $r=b$. 
    \item Tributaries of the FS given by Eq. (\ref{eq:tributaryFS}), they reach the DDR's body at $r=b$.
\end{enumerate}

In the following section, we describe how to compute each class of singular surface and its corresponding tributary trajectories.

\section{Motion strategies}
\label{sec:motion_strategies}

In this section, we compute the players' time-optimal motion strategies to reach their goals. We follow the method developed by R. Isaacs \cite{ISAACS-65} and some later extensions to compute singular surfaces \cite{LEWIN-12}. Isaacs' methodology relies on solving a set of boundary value problems involving a constrained optimization problem. The key idea is to integrate the motion equations backward from the game's ending configurations. This allows us to reconstruct the players' trajectories that lead to the terminal conditions. Additionally, the integration is performed so that the resulting trajectories minimize a cost function, in our case, the time to escape.  For a more detailed description of the procedure employed in this section, we refer the reader to \cite{ISAACS-65,BASAR-98,LEWIN-12}.

\subsection{Optimal controls}

To integrate the motion equations, first, we need to obtain the expressions of the optimal controls used by the players. This is done by constructing the Hamiltonian of the system. We have that
\begin{equation}
\label{eq:Hamiltonian}
H(\mathbf{x},\mathbf{\lambda},\mathbf{v},\mathbf{u}) = \mathbf{\lambda}^T \cdot f(\mathbf{x},\mathbf{v},\mathbf{u}) + L(\mathbf{x},\mathbf{v},\mathbf{u}),
\end{equation}
where $\mathbf{\lambda}^T$ are the costate variables and $L(\mathbf{x},\mathbf{v},\mathbf{u})$ is the cost function. For problems of minimium time $L(\mathbf{x},\mathbf{v},\mathbf{u}) = 1$. In our game, we have that
\begin{equation}
\label{eq:hamiltoniancartesian}
\begin{split}
&H(\mathbf{x}, \lambda, u_1, u_2, v_1, v_2) = \lambda_{x}\left( \frac{u_2 - u_1}{2b} \right)y  + \lambda_{x} v_1 \sin v_2 \\
&- \lambda_{y}\left( \frac{u_2 - u_1}{2b} \right) x - \lambda_{y}\left( \frac{u_1 + u_2}{2} \right) + \lambda_{y} v_1 \cos v_2 + 1,\\
\end{split}
\end{equation}
The optimal controls can be obtained from Eqs. (\ref{eq:hamiltoniancartesian}) and Pontryagin's Maximum Principle, which states that along optimal trajectories of the system
\begin{equation}
\label{eq:hamiltonianmax}
\begin{split}
&\min_v \max_u H(\mathbf{x}, \lambda, \mathbf{u}, \mathbf{v})=0,\\
&u^*=\arg \max_u H(\mathbf{x}, \lambda, \mathbf{u}, \mathbf{v}),\\
&v^*=\arg \min_v H(\mathbf{x}, \lambda, \mathbf{u}, \mathbf{v}),\\
\end{split}
\end{equation}
where $u^*$ and $v^*$ denote the optimal controls of the players. From Eqs. (\ref{eq:hamiltoniancartesian}) and (\ref{eq:hamiltonianmax}), we have that the DDR's optimal controls are
\begin{equation}
\label{eq:pursuerctrls}
\begin{split}
&u_1^* = V_r^{\max}\mbox{sgn}\left(\frac{-y\lambda_{x}}{b} + \frac{x\lambda_{y}}{b} - \lambda_{y} \right),  \\
&u_2^* = V_r^{\max}\mbox{sgn}\left( \frac{y\lambda_{x}}{b} - \frac{x\lambda_{y}}{b} - \lambda_{y} \right),
\end{split}
\end{equation}
and the OA's optimal controls are
\begin{equation}
\label{eq:evaderctrls}
v_1^* = V_a^{\max},\: \sin v_2^* = -\frac{\lambda_{x}}{\gamma},\: \cos v_2^* = -\frac{\lambda_{y}}{\gamma},
\end{equation}
where $\gamma = \sqrt{\lambda_{x}^2 + \lambda_{y}^2}$. 

The previous expressions for the players' optimal controls depend on knowing $\mathbf{\lambda}^T$ over time. These values are determined using the costate equations. Because the integration is conducted backward in time, beginning from the final conditions, we need the retro-time form of these equations. The retro-time is defined as $\tau=t_f-t$ where $t_f$ is the time of the game's termination. The {\em retro-time derivative} of a variable $x$ is denoted as $\mathring x$. The retro-time versions of the costate equations are
\begin{equation}
\mathring \lambda =\frac{\partial}{\partial x}H(\mathbf{x}, \lambda, u_1^*, u_2^*, v_1^*, v_2^*),
\end{equation}
for our problem
\begin{equation}
\label{eq:adjointeq}
\begin{split}
&\mathring \lambda_{x}= -\left( \frac{u_2^*  - u_1^*}{2b}\right) \lambda_{y}, \:\: \mathring \lambda_{y}= \left( \frac{u_2^*  - u_1^*}{2b} \right) \lambda_{x}.  \\
\end{split}
\end{equation}
From Eq. (\ref{eq:reducedsystem}), the retro-time version of the motion equations in the reduced space is
\begin{equation}
\label{eq:retroreducedsystem}
\begin{split}
&\mathring{x} =-\left(\frac{u_2  - u_1 }{2b}\right)y - v_1 \sin v_2, \\
&\mathring{y} =\left(\frac{u_2  - u_1 }{2b}\right)x + \left(\frac{u_1 + u_2 }{2}\right) - v_1 \cos v_2.
\end{split}
\end{equation}

The previous expressions were already obtained in \cite{RUIZ-22}, but we have included them to make the paper self-contained. 

\subsection{Terminal conditions}

As mentioned before, the procedure performs a backward integration of the motion equations starting from the game's ending configurations. Thus, we need to find the configurations in the playing space where the OA guarantees escaping from the detection region regardless of the DDR's control choice. This set is denoted as the usable part (UP). For this problem, the OA escapes from the DDR's detection region when the distance between them is greater than the radius $r_d$ despite any resistance of the DDR. In the reduced space, the terminal surface $\zeta$ can be parametrized by the angle $s$ (see Fig. \ref{fig:terminal}), which is the angle between the OA's position and the DDR's heading when the game ends (recall that in the reduced space all orientations are measured with respect to the positive $y$-axis in a clockwise sense).

At the game's end
\begin{equation}
\label{eq:terminalsurface}
x=r_d\sin s, \: y=r_d\cos s.
\end{equation} 
From \cite{ISAACS-65,BASAR-98}, the UP is described by
\begin{equation}
\label{eq:UP}
\mbox{UP}=\left\lbrace\mathbf{x} \in \zeta : \min_{v}\max_{u}\mathbf{n}\cdot f(\mathbf{x},u,v)<0\right\rbrace,
\end{equation}
where $\mathbf{n}$ is the normal vector from point $\mathbf{x}$ on $\zeta$ and extending into the playing space. From Eq. (\ref{eq:terminalsurface}), the  normal $\mathbf{n}$ to $\zeta$ in this game is given by
\begin{equation}
\label{eq:normal}
\mathbf{n}=\left[-\sin s \:\: -\cos s \right].
\end{equation}
Substituting Eq. (\ref{eq:normal}) and Eq. (\ref{eq:reducedsystem}) into Eq. (\ref{eq:UP}) we obtain
\begin{equation}
\label{eq:upfinal}
\mbox{UP}=\left\lbrace \max_{u_1,u_2} \bigg[-V_a^{\max} + \left(\frac{u_1  + u_2 }{2}\right) \cos s < 0\bigg]\right\rbrace.
\end{equation}

Different from \cite{RUIZ-22} where the DDR is faster than the OA and the UP depends on the value of $\rho_v$ (the players' ratio of velocities), the OA is faster than the DDR in our setup. Thus, from Eq. (\ref{eq:upfinal}), the UP corresponds to all configurations where the OA is located at a distance $r_d$. Note that once the evader has reached a distance $r_d$ to the DDR, it can increase it immediately by moving radially outwards to the DDR's position, regardless of the DDR's control choice. Conversely, the DDR moves at maximum translational speed at the game's end to counteract the OA's move. If $\cos s > 0$, we have that $u_1,u_2=V_r^{\max}$ and if $\cos s < 0$, we have that $u_1,u_2=-V_r^{\max}$.

\subsection{Primary solution}

In the following paragraphs, we compute the players' trajectories that lead directly to terminal conditions. They are known as the {\em primary solution}. At the UP, $x_0=r_d\sin s$ and $y_0=r_d\cos s$. Also, from the transversality conditions, we have that $\lambda_{x_0}=-\sin s$ and $\lambda_{y_0}=-\cos s$. As was presented in \cite{RUIZ-22}, given the DDR's moves at the maximum translational speed at the end of the game, then the solution of Eq. (\ref{eq:adjointeq}) is
\begin{equation}
\label{eq:adjointsolutions}
\lambda_{x}=-\sin s, \:\: \lambda_{y}=-\cos s.
\end{equation}
Integrating the retro-time version of the motion equations in Eq. (\ref{eq:retroreducedsystem}), we have that
\begin{equation}
\label{eq:straightline}
\begin{split}
&x = -\tau V_a^{\max} \sin s  + r_d \sin s,\\
&y = \tau(-V_a^{\max} \cos s \pm V_r^{\max}) + r_d \cos s,\\ 
\end{split}
\end{equation}
the sign $+$ is taken if the DDR moves forward in the realistic space and the sign $-$ if it moves backward. 

\subsection{Transition surface}

Eqs. (\ref{eq:adjointsolutions}) and (\ref{eq:straightline}) remain valid as long as the players do not switch controls. In this game, after a retro-time interval, the DDR switches control and begins rotating in place in the realistic space. The set of configurations where a control variable changes value is called a {\em transition surface} (TS).

The time $\tau_s$ when the DDR switches controls is computed substituting Eq. (\ref{eq:adjointsolutions}) and Eq. (\ref{eq:straightline}) into Eq. (\ref{eq:pursuerctrls}), and verifying which one of the resulting expressions change its sign first. We found that for $s\in [0, \frac{\pi}{2}]$, $u_2^*$ switches first from $-V_r^{\max}$ to $V_r^{\max}$ (i.e., the DDR's controls after the switch are $u_1^*=V_r^{\max}$ and $u_2^*=-V_r^{\max}$) and it does it at 
\begin{equation}
\label{eq:firstswitch}
\tau_s = \frac{b \cot s}{V_r^{\max}}.
\end{equation}
A similar expression can be obtained for the remaining quadrants.

\subsubsection{Tributary trajectories of the TS}

At $\tau_s$, a new integration of Eq. (\ref{eq:adjointeq}) and Eq. (\ref{eq:reducedsystem}) is required. This integration takes as initial conditions the values of $\lambda_{x}$, $\lambda_{y}$, $x$, and $y$ at $\tau_s$. Those values are denoted as $\lambda_{x_s}$, $\lambda_{y_s}$, $x_s$ and $y_s$. The solution of the Eq. (\ref{eq:adjointeq}) is
\begin{equation}
\label{eq:adjointsolutions2}
\begin{split}
\lambda_{x} = -\sin\left[s - \left(\frac{u_2^*-u_1^*}{2b}\right)\tau_1\right], \\
\lambda_{y} = -\cos\left[s - \left(\frac{u_2^*-u_1^*}{2b}\right)\tau_1\right],
\end{split}
\end{equation}
where $\tau_1=\tau - \tau_s$. The corresponding trajectories in the reduced are given by
\begin{equation}
\label{eq:rotation}
\begin{split}
&x = -y_s\sin\left[\left(\frac{u_2^*-u_1^*}{2b}\right)\tau_1\right]+x_s\cos\left[\left(\frac{u_2^*-u_1^*}{2b}\right)\tau_1 \right]\\
&-\tau_1 V_a^{\max}\sin\left[s-\left(\frac{u_2^*-u_1^*}{2b}\right)\tau_1\right],\\
&y = x_s\sin\left[\left(\frac{u_2^*-u_1^*}{2b}\right)\tau_1\right]+y_s\cos\left[\left(\frac{u_2^*-u_1^*}{2b}\right)\tau_1 \right]\\
&-\tau_1 V_a^{\max}\cos\left[s-\left(\frac{u_2^*-u_1^*}{2b}\right)\tau_1\right].
\end{split}
\end{equation}

\subsection{Discussion about the primary solution, the TS and its tributaries}

One interesting behavior of the primary solution is that a subset of it contains trajectories that intersect the $y$-axis. To compute them, we make the first expression in Eq. (\ref{eq:straightline}) equal to zero ($x=0$), and we obtain $\tau_c=\frac{r_d}{V^{\max}_a}$, the time that took them to reach the $y$-axis. From the previous result, we can observe that all trajectories of the primary solution intersect the $y$-axis simultaneously. Substituting $\tau_c$ into the second expression of Eq. (\ref{eq:straightline}), we have that $y_c=\frac{r_d}{\rho_v}$, i.e., all trajectories intersecting the $y$-axis do it at the same point. The last trajectory of the primary solution reaching $y_c$ does it exactly at the moment the DDR switches controls and is given by $s_c=\arctan{(\rho_v \rho_d)}$. The previous analysis indicates that at point $(0,y_c)$, the evader's control $v_2^*$ is not unique, and it has the opportunity to select a value of $v^*_2\in[0,s_c]$, each one resulting in the same outcome. Once the evader has picked $v_2^*$ and has departed from $(0,y_c)$, it has to stick to its choice until the game's end.

We discovered that the tributary trajectories of the TS in quadrant I, $\phi \in [0,\frac{\pi}{2})$, intersect with the ones in quadrant II, $\phi \in (\frac{\pi}{2}, \pi]$, at the positive $x$-axis. The configurations where the intersection occurs define a Dispersal Surface (DS). On that surface, the players have two choices for their optimal controls, resulting in trajectories having the same cost, but each control is linked to the other player's selection. The DDR can rotate clockwise or counterclockwise, but its choice must correspond to the evader's choice of moving either toward quadrant I or II. Once the players leave the DS, they must stick to their controls until they reach the TS. An equivalent DS is produced by the intersection of the tributary trajectories of the TS in quadrant III, $\phi \in [\pi, \frac{3\pi}{2})$,  and the ones in quadrant IV, $\phi \in (\frac{3\pi}{2}, 2\pi]$, at the negative $x$-axis.

Another DS may appear for some values of $\rho_v$, which is created by tributary trajectories of the TS in quadrant I,  $\phi \in [0,\frac{\pi}{2})$, intersecting with the ones in quadrant IV, $\phi \in (\frac{3\pi}{2}, 2\pi]$, at the positive $y$-axis. In this case, a similar DS arises at the negative $y$-axis produced by the intersection of tributary trajectories of the TS in quadrant II, $\phi \in (\frac{\pi}{2}, \pi]$, and quadrant III, $\phi \in [\pi, \frac{3\pi}{2})$.

In this game, we encountered that for some values of $\rho_d$ and $\rho_v$, the primary solution and the tributaries reaching the TS are not enough to cover the playing space. Since the OA is faster than the DDR and it can always escape from the detection region, this suggests that additional singular surfaces and their corresponding tributaries may exist. In the following paragraphs, we proceed to construct them.

\subsection{Universal surface}

As was mentioned in the previous section, a portion of the primary solution trajectories intersects the $y$-axis at $y_c=\frac{r_d}{\rho_v}$ and leaves a void region in the playing space. As it has been observed in other pursuit-evasion games \cite{LEWIN-12}, this behavior indicates the existence of a singular surface. In this game, we found that a portion of the $y$-axis corresponds to a Universal Surface (US). In the US, one of the players dominates and can force the other player to remain in that trajectory; otherwise, the non-dominant player will benefit the other player. The DDR has a rotational advantage in the US and can maintain its heading aligned with the OA's position. Thus, the DDR translates towards the OA while the evader moves away from it radially. The trajectory of the system in the reduced space is given by
\begin{equation}
\label{eq:trajectoryUS}
\begin{split}
x &= 0, \\
y &= (-V_a^{\max}\pm V_r^{\max})\tau_2 + y_c,
\end{split}
\end{equation}
where $\tau_2=\tau-\tau_c$. The sign $+$ is taken when the DDR moves forward in the realistic space, and $-$ if it moves backward.

We noticed that, depending on the value of $\rho_v$, the segment between $(0,y_c)$ and $(0,b)$ corresponds only to the US, but in other cases, there is a Dispersal Surface (DS) between the US and the point $(0,b)$, which is generated by the tributary trajectories of the US meeting again the $y$-axis. Unfortunately, we did not find analytically the point in the $y$-axis where the US ends and the DS begins.

\subsubsection{Tributary trajectories}

To find the tributary trajectories of the US, we need to determine which player switches control when the system arrives in the US. In particular, the trajectory of the primary solution, reaching $y_c$ exactly when the DDR switches controls, provides a hit of the behavior of the tributary trajectories in the US. Since that trajectory and the first one departing from the US must join smoothly, we have that the tributary trajectories of the US correspond to the DDR performing a rotation in place at maximum speed while the evader follows a straight-line trajectory. For $s\in [0, \frac{\pi}{2})$, the DDR's controls on the tributary trajectories are $u_1^*=V_r^{\max}$ and $u_2^*=-V_r^{\max}$. Considering that $u_2$ switches when the system leaves the US and starts following a tributary trajectory in retro-time, from Eq. (\ref{eq:pursuerctrls}) we have that on that moment $\frac{y\lambda_{x}}{b} - \frac{x\lambda_{y}}{b} - \lambda_{y}=0$. Denoting $(0,y_u)$ to the point in the US when the tributary trajectory meets the US, we get that $\frac{\lambda_x}{\lambda_y}=\frac{b}{y_u}$. From Eq. (\ref{eq:evaderctrls}) and the previous expression, we found that $v_2^*=\frac{b}{y_u}$ on the tributary trajectory. Integrating Eq. (\ref{eq:retroreducedsystem}) considering the players' optimal controls, we have that
\begin{equation}
\label{eq:rotationUS}
\begin{split}
&x = -y_u\sin\left[\left(\frac{u_2^*-u_1^*}{2b}\right)\tau_{3}\right]\\
&-\tau_3 V_a^{\max}\sin\left[v_2^*-\left(\frac{u_2^*-u_1^*}{2b}\right)\tau_3\right],\\
&y = y_u\cos\left[\left(\frac{u_2^*-u_1^*}{2b}\right)\tau_3\right]\\
&-\tau_3 V_a^{\max}\cos\left[v_2^*-\left(\frac{u_2^*-u_1^*}{2b}\right)\tau_3\right],
\end{split}
\end{equation}
where $\tau_3=\tau-(\tau_c+\frac{|y_u-y_c|}{V_a^{\max}-V_r^{\max}}$). 

Similarly to the tributary trajectories of the TS, we discovered that in some cases, depending on the value of $\rho_v$, a portion of the tributary trajectories of the US in Quadrant I intersect with the ones in Quadrant II, at the positive $x$-axis. These configurations define a Dispersal Surface.

\subsection{Focal surface}

We found that for some values of $\rho_v$, one of the tributary trajectories of the TS in quadrant I meets tangentially with the positive $x$-axis. An identical behavior occurs with the symmetric tributary trajectories of the TS in quadrant II. A void region appears between the point where both trajectories meet tangentially and the robot's body. Similar to the US, this indicates a singular surface's existence. In this case, the point $(x_f,0)$ where both trajectories meet corresponds to the beginning of a Focal Surface (FS). We did not succeed in finding this point analytically since the tributaries trajectories of the TS are given by transcendental equation. However, it can be approximated using numerical analysis. 

In the FS, one player dominates the other and forces it to remain in that trajectory; particularly, the system stays on a portion of the $x$-axis. In this case, the OA can maintain the DDR in the perpendicular condition, i.e., countering the DDR rotation and simultaneously increasing the distance from it. Since the evader makes $\dot y = 0$, then from Eq. (\ref{eq:retroreducedsystem}), we have that $v^*_2 = \arccos{\left(\frac{-\rho_v x}{b}\right)}$. The DDR performs a rotation in place at maximum speed simultaneously. We get that the retro-time motion equations in Eq. (\ref{eq:retroreducedsystem}) take the form 
\begin{equation}
\label{eq:retroreducedsystemFS}
\begin{split}
&\mathring{x} = - V_a^{\max} \sqrt{1-\left(\frac{-\rho_v x}{b}\right)^2}  ,\\
&\mathring{y} = 0.
\end{split}
\end{equation}
Integrating Eq. (\ref{eq:retroreducedsystemFS}) from $(x_f,0)$, we get that
\begin{equation}
\begin{split}
x &= \left(\frac{b}{\rho_v}\right)\sin \left(\frac{\rho_v V_a^{\max} \tau_4}{b}  - \arcsin{\left( - \frac{ x_f\rho_v}{b}\right)}\right), \\
y &= 0, \\
\end{split}    
\end{equation}
where $\tau_4=\tau -\tau_f$, and $\tau_f$ denotes the time when the tributary trajectory of the TS reaches the point~$(x_f,0)$.

\subsubsection{Tributary trajectories}

To compute the tributary trajectories reaching the FS, we need to find the player's optimal controls when they arrive at the FS. We know that when a tributary trajectory reaches the FS, it does it tangentially. Let $(x_t,0)$ denote the point when that occurs. From the tangential condition, we have that $
v_2^* = \arccos{\left( \frac{-\rho_v x_t}{b} \right)}
$. 
Integrating Eq. (\ref{eq:retroreducedsystem}) considering the previous value of $v_2^*$ and that the DDR performs a rotation in place at maximum speed, we have that
\begin{equation}
\label{eq:tributaryFS}
\begin{split}
&x = x_f\cos\left[\left(\frac{u_2^*-u_1^*}{2b}\right)\tau_5\right]\\
&-\tau_5 V_a^{\max}\sin\left[v_2^*-\left(\frac{u_2^*-u_1^*}{2b}\right)\tau_5\right],\\
&y = x_f\sin\left[\left(\frac{u_2^*-u_1^*}{2b}\right)\tau_5\right]\\
&-\tau_5 V_a^{\max}\cos\left[v_2^*-\left(\frac{u_2^*-u_1^*}{2b}\right)\tau_5\right],
\end{split}
\end{equation}
where $\tau_5=\tau-(\tau_f+\tau_{t})$, and $\tau_t$ denotes the time the system travels over the FS, from point $(x_f,0)$ to point $(x_t,0)$.

We must mention that when the FS appears in the positive $x$-axis, an analogous one emerges in the negative $x$-axis. That singular surface and its tributary trajectories can be obtained using similar reasoning.

Another important behavior is that in some cases, again, depending on the value of $\rho_v$, none of the tributary trajectories of the TS in quadrants I and II reach the positive $x$-axis tangentially. However, the FS may appear and be generated by the tributary trajectories of the US in quadrants I and II if two of those trajectories reach tangentially the positive $x$-axis.

\begin{figure*}[t!]
    \centering
    \input{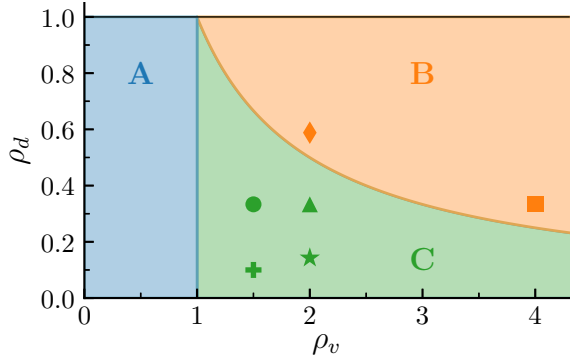}
    \caption{Diagram of $\rho_v$ vs $\rho_d$.
    A:~Case of a slower evader $V_a^{\max}<V_r^{\max}$ studied in \cite{RUIZ-22}.
    B:~No critical point appears.
    C:~Critical point $(0,y_c)$ appears.
    Symbols correspond to the cases shown in 
    ($\star$)~Fig.~\ref{fig:partition}, 
    ($\bullet$)~Fig.~\ref{fig:VaryingVa_a}, 
    ($\blacksquare$)~Fig.~\ref{fig:VaryingVa_b}, 
    ($\blacklozenge$)~Fig.~\ref{fig:VaryingRd_a}, 
    ($\blacktriangle$)~Fig.~\ref{fig:VaryingRd_b}, and 
    (\textbf{+})~Fig.~\ref{fig:VaryingRd_c}. 
    }
    \label{fig:phasediagram}
\end{figure*}

\section{Characterization of the solution based on $\rho_v$ and $\rho_d$}
\label{sec:partition}

For different values of the parameters $V_a^{\max}$ and $r_d$, some of the singular surfaces may appear, and some of them may not. To see this in more detail, we characterize different regions in a diagram of $\rho_v$ vs $\rho_d$ as shown in~Fig.~\ref{fig:phasediagram}. 
\begin{enumerate}[label=\Alph*.,itemsep=0pt]
    \item For $\rho_v<1$, the partition of the playing space corresponds to the case studied in \cite{RUIZ-22}, when the OA is slower than the DDR. %
    Namely, a state-constrained (SC) surface at $r=r_d$ and a barrier appear. %
    \item For $\rho_v>1$ and $\rho_v\rho_d>b$, the TS intersects with the DDR's body at~$y<b$, the critical point~$(0,y_c)$ does not appear, and the only singular surface is a horizontal DS reached by the tributaries of the TS. %
    \item For $\rho_v>1$ and $\rho_v\rho_d<b$, the primary trajectories with $s<\arctan(\rho_v\rho_d)$ converge at the critical point~$(0,y_c)$ and additional singular surfaces like the US and the FS may appear depending on the specific values of $V_a^{\max}$ and $r_d$.
\end{enumerate}

\begin{figure*}[t!]
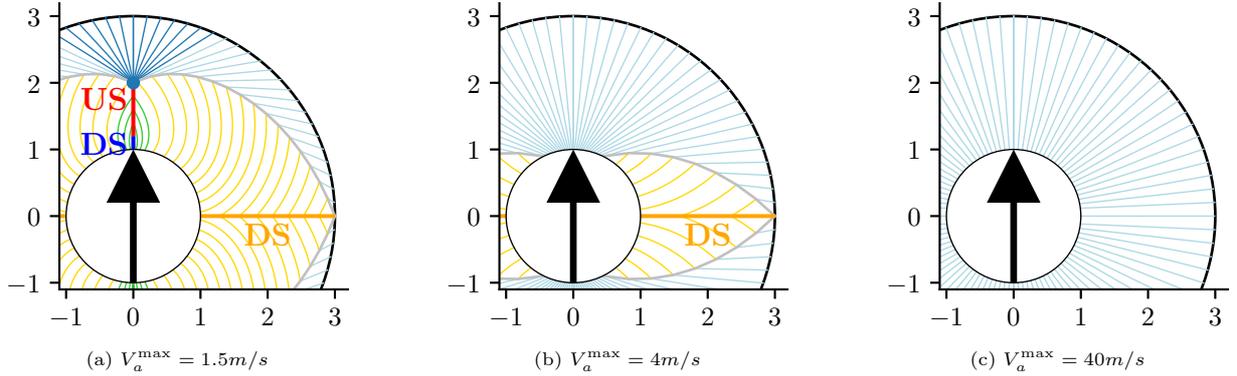

    \centering
    \subfloat[$V_a^{\max}=1.5m/s$]{
        \input{fig4_VaryingVa_a.pgf}
        \label{fig:VaryingVa_a}
    }
    \hspace{0.05\linewidth}
    \subfloat[$V_a^{\max}=4m/s$]{
        \input{fig4_VaryingVa_b.pgf}
        \label{fig:VaryingVa_b}
    }
    \hspace{0.05\linewidth}
    \subfloat[$V_a^{\max}=40m/s$]{
        \input{fig4_VaryingVa_c.pgf}
        \label{fig:VaryingVa_c}
    }
    \caption{\label{fig:VaryingVa}
        Partition of the playing space for fixed DDR's detection range $r_d=3m$ and different values of the OA's speed $V_a^{\max}$. 
    }
\end{figure*}
We investigate the singular surfaces that can appear as we increase the speed $V_a^{\max}$ of the evader at fixed detection radius $r_d=3m$, see Fig. \ref{fig:VaryingVa}. For a speed of $V_a^{\max}=1.5m/s$ (only slightly larger than that of the DDR, $V_r^{\max}=1m/s$), we observe the emergence of the critical point $(0,y_c)$, as well as singular surfaces like the US at $x=0$, and both horizontal and vertical DS, see Fig. \ref{fig:VaryingVa_a}. Note that in this case, the FS does not appear, as here, tributaries of the TS intersect transversally with the horizontal DS. For a larger speed $V_a^{\max}=4m/s$, the critical point $(0,y_c)$ does not show anymore, and the TS intersects the DDR at $y<b$, the only singular surface that appears is the DS at $y=0$, see Fig. \ref{fig:VaryingVa_b}. 
When the OA's speed is very large, $V_a^{\max}=40m/s$, the time-optimal trajectories correspond to radial straight lines, see Fig. \ref{fig:VaryingVa_c}. This result agrees with the intuition that the OA's strategy to escape the detection range is to move radially outward from the DDR, which we have found to be optimal only in the limit $V_a^{\max}\gg{V}_r^{\max}$.

\begin{figure*}[t!]
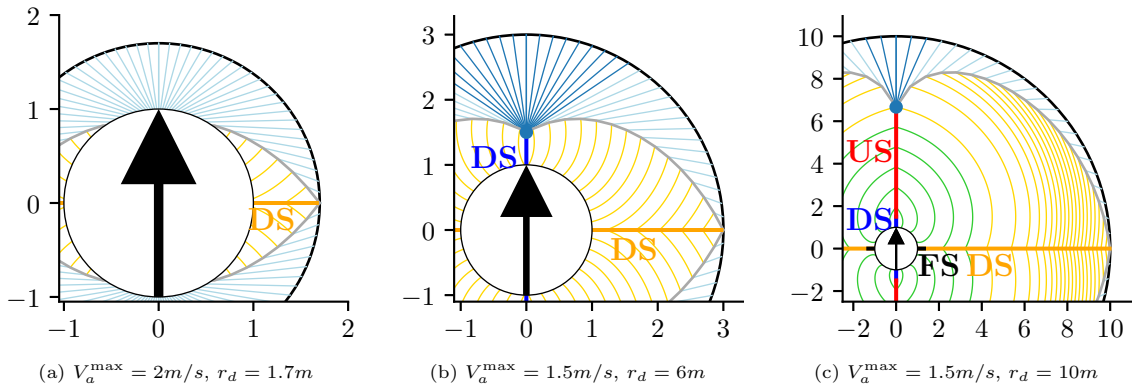

    \centering
    \subfloat[$V_a^{\max}=2m/s$, $r_d=1.7m$]{
        \input{fig5_VaryingRd_a.pgf}%
        \label{fig:VaryingRd_a}%
    }%
    \hspace{0.03\linewidth}%
    \subfloat[$V_a^{\max}=1.5m/s$, $r_d=6m$]{
        \input{fig5_VaryingRd_b.pgf}%
        \label{fig:VaryingRd_b}%
    }%
    \hspace{0.03\linewidth}%
    \subfloat[$V_a^{\max}=1.5m/s$, $r_d=10m$]{
        \input{fig5_VaryingRd_c.pgf}%
        \label{fig:VaryingRd_c}%
    }%
    \caption{\label{fig:VaryingRd}%
        Partition of the playing space for different values of the detection radius~$r_d$. %
    }
    \label{fig:varyingRd}
\end{figure*}
Different cases of singular surfaces appearing are also found as we vary the detection range~$r_d$, see Fig. \ref{fig:VaryingRd}. For $V_a^{\max}=2m/s$ and detection radius $r_d=1.7m$, we again see the TS intersects with the DDR's body, and the horizontal DS is the only singular surface that appears, see Fig. \ref{fig:VaryingRd_a}. For $V_a^{\max}=1.5m/s$ and a larger detection radius $r_d=6m$, the vertical DS and the critical point $(0,y_c)$ appear, see Fig.~\ref{fig:VaryingRd_b}. The US is not present in this case, as tributaries from the TS transversely reach the vertical DS. For an even bigger detection radius $r_d=10m$, the US emerges, as well as the FS, which now appears due to a tributary trajectory of the US that is tangential to the horizontal DS at $y=0$, see Fig. \ref{fig:VaryingRd_c}.

\section{Simulations}
\label{sec:simulations}

To illustrate the players' time-optimal motion strategies in the realistic space, we perform simulations of the differential game by numerically integrating Eqs. (\ref{eq:DDRmotion}) and (\ref{eq:OAmotion}). For this, we consider the optimal controls $u_1^*$, $u_2^*$, $v_1^*$, and $v_2^*$ that we obtained in the reduced space by employing Isaacs' method as described in Sec.~\ref{sec:motion_strategies}. The time-optimal motion strategies presented in \cite{RUIZ-22} correspond to the case $\rho_v<1$, where the game takes place in regions similar to I and II (as described in Sec. \ref{sec:partition}). This section focuses on cases when the game occurs in regions III and IV, where the US and the FS appear. Note that trajectories in these regions correspond to motion strategies specific to the case of a faster OA.

\begin{figure*}[t!]
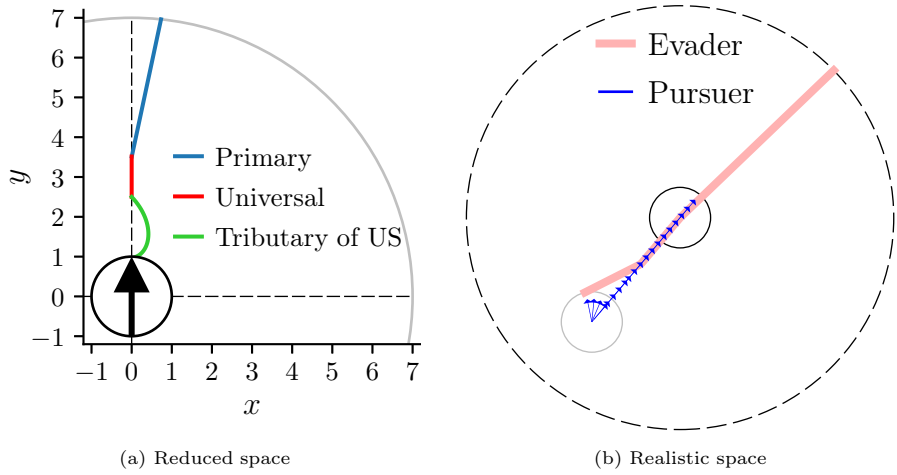

    \centering
    \subfloat[Reduced space]{
    \input{fig6_Simulation1_a.pgf}%
    \label{fig:simulation1_reducedspace}
    }%
    \hspace*{0.02\linewidth}%
    \subfloat[Realistic space]{
    \input{fig6_Simulation1_b.pgf}
    \label{fig:simulation1_realisticspace}
    }%
    \caption{%
        Time-optimal motion strategy traveling a primary trajectory, a portion of the US, and a tributary trajectory.
    }
\end{figure*}

First, we consider the case when the OA is initially in contact with the DDR at~$(x,y)\approx(0,b)$. Note the initial position is not exactly at $x=0$, but it is slightly displaced to the right; see Fig.~\ref{fig:simulation1_reducedspace}. For parameters $r_d=7m$, $V_r^\mathrm{max}=1m/s$, and $V_a^\mathrm{max}=2m/s$, the time-optimal motion strategy in the reduced space corresponds to moving first along a tributary of the US. In the realistic space, this equals to a rotation in place for the DDR, while the OA attempts to escape by following a straight-line trajectory, see Fig.~\ref{fig:simulation1_realisticspace}. Eventually, the OA reaches the Universal Surface, and the DDR performs a straight-line pursuit. Note that while the OA is ensured to escape due to its speed advantage, for it is not convenient to remain aligned with the DDR's direction of motion, as it would take more time to escape. However, as the OA distance from the DDR becomes larger, it becomes easier for the DDR to force and maintain alignment with $x=0$, i.e., the DDR has a rotation advantage. In the end, the OA reaches the critical point at $(x,y)=(0,b/\rho_v)$ where it has the freedom of choosing a control $v_2^*\in[-s_c,s_c]$. Once the OA has made its decision, it has to stick to it, resulting in a straight-line trajectory that reaches the terminal surface. It is important to mention that all straight-line trajectories produced by $v_2^*\in[-s_c,s_c]$ have the same cost. Some snapshots of the simulation are shown in Fig.~\ref{fig:simulation1_snapshots}.

\begin{figure*}[t!]
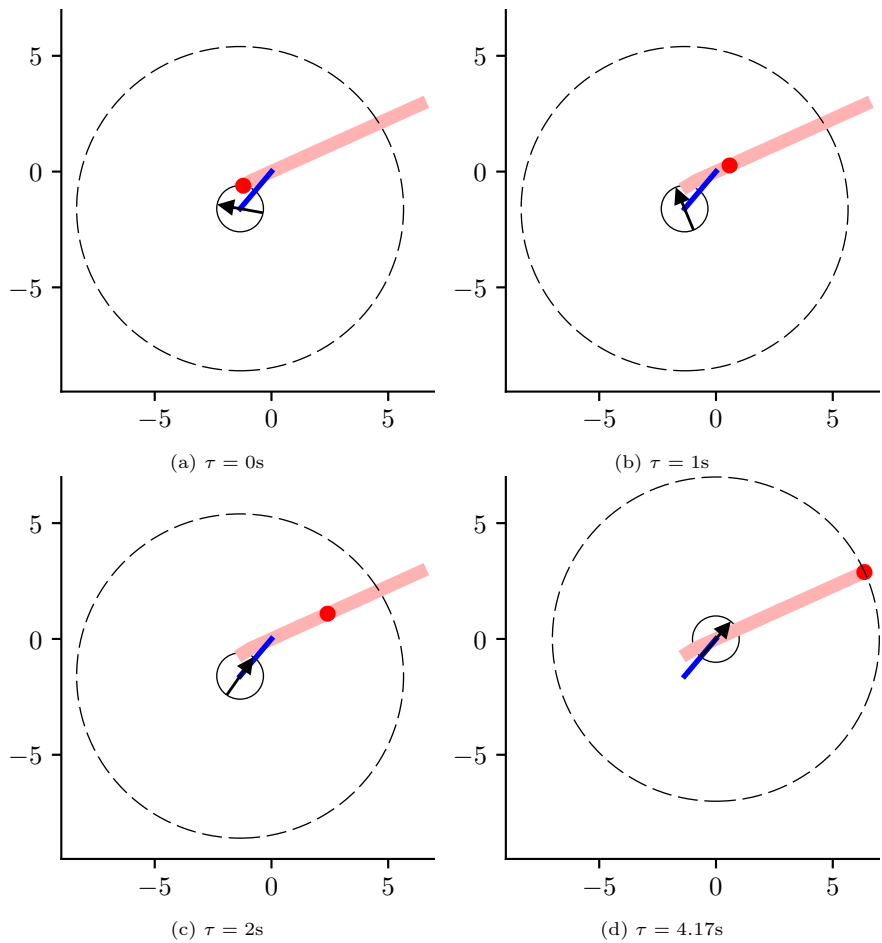

    \centering
    \subfloat[$\tau =0\mathrm{s}$]{
        \input{fig9_Snapshots_a.pgf} 
    }%
    \subfloat[$\tau =1\mathrm{s}$]{
        \input{fig9_Snapshots_b.pgf} 
    }\\
    \subfloat[$\tau =2\mathrm{s}$]{
        \input{fig9_Snapshots_c.pgf} 
    }%
    \subfloat[$\tau =4.17\mathrm{s}$]{
        \input{fig9_Snapshots_d.pgf} 
    }%
    \caption{%
        Simulation snapshots.%
        \label{fig:simulation2_snapshots}%
    }
\end{figure*}

Second, we consider the case when the OA is initially located at $(x,y)\approx(b,0)$. The initial position is not exactly at $y=0$, but it is slightly displaced upwards; see Fig. \ref{fig:simulation2_reducedspace}. The strategy here is to first move along a tributary trajectory of the FS, where the DDR rotates in place, and the OA attempts to stay on a lateral position with respect to the DDR's motion direction, see Fig. \ref{fig:simulation2_realisticspace}. In this situation, the DDR has a significant disadvantage, as its only option is to keep rotating in place at maximum speed, trying to align with the OA's position. When the OA reaches the focal surface at $y=0$, it slightly switches its direction of motion to keep counteracting the DDR's motion. After some time, the system follows a tributary trajectory of the TS, corresponding again to a DDR's rotation in place at a maximum speed and a straight-line motion for the OA. Finally, along the primary trajectory, the DDR is forced to perform a straight-line pursuit (it is the best it can do even when the DDR's heading is not perfectly aligned with the OA's position), whereas the OA keeps the same straight-line trajectory until it reaches the terminal surface. Some snapshots of the second simulation are shown in Fig. \ref{fig:simulation2_snapshots}.

\begin{figure*}[t!]
    \centering
    \subfloat[$\tau =0\mathrm{s}$]{
        \input{fig7_Snapshots_a.pgf} 
    }%
    \subfloat[$\tau =1\mathrm{s}$]{
        \input{fig7_Snapshots_b.pgf} 
    }\\
    \subfloat[$\tau =3\mathrm{s}$]{
        \input{fig7_Snapshots_c.pgf} 
    }%
    \subfloat[$\tau =5.5\mathrm{s}$]{
        \input{fig7_Snapshots_d.pgf} 
    }%
    \caption{%
        Simulation snapshots.%
        \label{fig:simulation1_snapshots}%
    }
\end{figure*}

\begin{figure*}[t!]
    \centering
    \subfloat[Reduced space]{%
        \input{fig8_Simulation2_a.pgf}%
        \label{fig:simulation2_reducedspace}%
    }%
    \hspace*{0.02\linewidth}%
\subfloat[Realistic space]{%
        \input{fig8_Simulation2_b.pgf}%
        \label{fig:simulation2_realisticspace}%
    }
    \caption{%
        Time-optimal motion strategy traversing a primary trajectory, a portion of the FS, and a tributary trajectory.
    }
\end{figure*}

\section{Conclusions and future work}
\label{sec:conclusions}

In this work, we studied the problem of maintaining a faster OA inside the bounded detection region of a DDR for as much time as possible. We formulated the problem as a zero-sum differential game, and we computed the time-optimal motion strategies of the players to achieve their goals. Given the OA's speed advantage, a winning strategy for the OA is always moving radially outwards to the DDR's position. We showed that even though the previous strategy could be optimal in some cases, more complex time-optimal motion strategies emerge based on the players' speed ratio $\rho_v$. In particular, we exhibit that four classes of singular surfaces may appear in this game: Dispersal, Transition, Universal, and Focal surfaces. Additionally, some numerical simulations are included to illustrate the time-optimal motion strategies of the players in the reduced and realistic spaces. 

Three interesting research avenues to continue this work are as follows. The first implies dealing with detection regions with range and angular bounds. In this case, a key question to be answered is when the escape occurs by reaching the range bounds and when it is attained by achieving the angular bounds. The second is developing cooperative motion strategies in which two or more DDR pursuers work together to maintain surveillance of the OA. Another more challenging task is creating a strategy for the more general case where the players move in environments with obstacles. This implies dealing with fundamental questions in pursuit-evasion games, like how to model the obstacles and how to extend or develop new techniques that provide theoretical guarantees in those environments.

\section{Acknowledgments}

CONAHCYT supported this work under Grant A1-S-21934.

\bibliographystyle{unsrt}        
\bibliography{main}           

\end{document}